\documentclass[letterpaper, 10 pt, conference]{ieeeconf}  

\IEEEoverridecommandlockouts                              

\usepackage{hyperref} 
\usepackage{pifont}
\usepackage{amsmath}
\usepackage{adjustbox} 
\usepackage{multirow}
\usepackage{multicol}
\usepackage{graphicx}
\usepackage{subfig}
\usepackage{booktabs}
\usepackage{times}
\usepackage{textcomp}
\usepackage{color}
\usepackage{cite}
\usepackage{breakcites}
\usepackage{dblfloatfix}

\title{\LARGE \bf
``What's This?'' - Learning to Segment Unknown Objects from Manipulation Sequences
}

\author{Wout Boerdijk$^{1}$ \and Martin Sundermeyer$^{1}$ \and Maximilian Durner$^{1}$ \and Rudolph Triebel$^{1,2}$
\thanks{$^{1}$Institute of Robotics and Mechatronics, German Aerospace Center (DLR),
        82234 Wessling, Germany
        {\tt\small <first>.<second>@dlr.de}}%
\thanks{$^{2}$Department of Computer Science, Technical University of Munich (TUM),
        85748 Garching, Germany}%
}

\begin{document}

\maketitle
\thispagestyle{empty}
\pagestyle{empty}

\begin{abstract}
	
We present a novel framework for self-supervised grasped object segmentation 
with a robotic manipulator.
%
Our method successively learns an agnostic foreground segmentation
followed by a distinction between manipulator and object solely by
observing the motion between consecutive RGB frames.
%
In contrast to previous approaches, we propose a single, end-to-end
trainable architecture which jointly incorporates motion cues and
semantic knowledge.
Furthermore, while the motion of the manipulator and the object are
substantial cues for our algorithm, we present means to robustly deal
with distraction objects moving in the background, as well as with
completely static scenes.  
Our method
neither depends on any visual registration of a kinematic robot or 3D object models, nor on precise
hand-eye calibration or any additional sensor data.
By extensive experimental evaluation we demonstrate the superiority of
our framework and provide detailed insights on its capability of
dealing with the aforementioned extreme cases of motion. 
We also show that training a semantic segmentation network with the automatically labeled data achieves results on par with manually annotated training data.
Code and pretrained model are available at \href{https://github.com/DLR-RM/DistinctNet}{https://github.com/DLR-RM/DistinctNet}.


\end{abstract}

\section{INTRODUCTION}


One of the key functionalities for any robotic manipulation system is
the ability to accurately determine the location of the object(s) to
be manipulated. To do this, most current systems employ computer
vision techniques, where the aim is to find the pixels that correspond
to the target object within a camera image taken from the scene. This
is called \emph{object segmentation}, and it is an important
prerequisite for a large number of downstream tasks such as object
pose estimation, 3D reconstruction, or grasp detection. In general,
object segmentation can be treated in two different ways, depending on
whether a representation of the object is given beforehand, for example in
form of a 3D model. If such a model exists, the segmentation problem
is significantly easier, because the model can be used to guide the
search, e.g. by training an object detector. Unfortuntaley, it is also
the less common case, because 3D models are often not available.

\begin{figure}[!t]
	\centering \captionsetup{width=\linewidth} \includegraphics[width=0.9\linewidth]{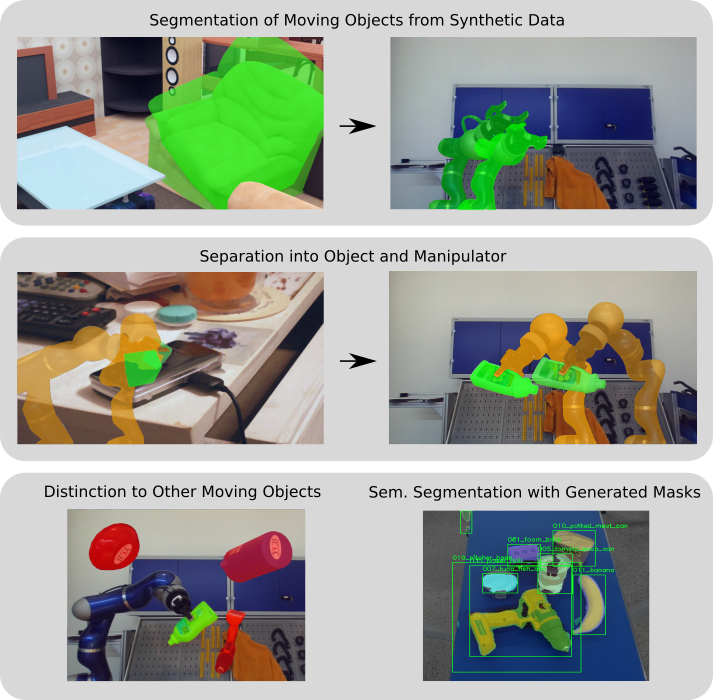} \caption{Our
	approach first learns to segment objects that move in the
	foreground (top). Then, we use fine-tuning to learn a
	segmentation between the manipulator and the unknown object
	(center). Furthermore, our model can distinguish moving
	background objects (bottom left) and support
	downstream tasks such as semantic segmentation (bottom right).  \label{fig:cover}}
\end{figure}%

Therefore, in this work we address segmentation of unknown objects,
where the only cue we rely on is the change of pixels that occurs when
moving the object in front of the camera. This can either be done by
pushing the object on a
table \cite{Fitzpatrick2003GroundingVT,schiebener_physical_2014,pathak_learning_2018,deng2020self}
or by holding it with a gripper and moving it with the manipulator. We
follow the latter approach, because then the object can be seen from
all sides, including from the bottom. We also
employ \emph{self-supervised} object segmentation, i.e. for segmentation we neither
require a kinematic model of the manipulator nor an intrinsic or
extrinsic calibration of the camera. This makes our approach
particularly versatile and easy to use. Also, in contrast to existing
works, we do not perform a two-step process where first a representation of the
manipulator is learned, before the actual object segmentation can be
done. Instead, we propose an end-to-end learning scheme where the
segmentation of both the manipulator and the object are learned
simultaneously. This has the major advantage that information can be
shared between the two tasks, and hyperparameter tuning or post
processing are not required. Furthermore, 
our approach is able to distinguish between objects
that are actively moved by the manipulator and those that move in the
background (see Fig.~\ref{fig:cover}). Finally, our architecture
provides means to segment the object even in cases where no motion is
observed for a small number of frames. As we will show experimentally,
our approach outperforms previous self-supervised methods, and it even
competes with techniques that rely on hand-eye calibration and additional depth data.

\section{Related Work}

We briefly discuss existing work on self-supervised object segmentation.
Additionally, we review methods that reason about correspondences between a pair of images.
Relevant to this work are also approaches which combine motion and semantic segmentation.

\subsection{Self-Supervised Grasped Object Segmentation}

A robot arm has initially been utilized for object segmentation by inducing motion in a static scene (e.g.~\cite{Fitzpatrick2003GroundingVT,schiebener_physical_2014,pathak_learning_2018, deng2020self}).
To have more control over the object motion, grasping the respective item is a reasonable consequential step - however, it additionally requires distinguishing the object from the manipulator.
This can be addressed with the help of a 3D robot model which allows to isolate the object~\cite{krainin_manipulator_nodate,welke_autonomous_2010}. 
Browatzki~et~al.~\cite{browatzki_active_2012} employ a perception-driven object recognition framework where the robot itself controls the object views, and segment the item area with a GMM.
Instead, \cite{rocha_self-supervised_2019} propose a self-supervised approach for pixel-wise robot recognition:
By projecting a robot model onto the image plane and simultaneously optimizing a GrabCut-based cost function, segmentation labels of the robot arm are obtained.
Florence~et~al.~\cite{florence_robot-supervised_2020} build upon this framework and extend it to grasped object segmentation:
They segment the foreground by projecting link positions of the robot into the camera frame which aids a graph-based depth segmentation followed by an additional refinement in RGB space.
With this approach they collect segmentation masks of the manipulator to learn a robot arm representation. 
During inference they obtain a joint robot-object mask with the same foreground segmentation, and subtract the manipulator with help of the robot arm segmentation network.
In our previous work~\cite{boerdijk_self-supervised_2020} we alleviate the constraints of the aforementioned methods in form of precise camera calibration and additional sensor modalities by observing motion in a static scene, and learn a manipulator representation from thresholded optical flow only.
Excluding the manipulator estimation from the joint object-manipulator mask - again derived by motion - yields the final object mask.

Yet, optical flow incorporates little semantic knowledge which requires additional post-processing. 
Moreover, none of the above approaches utilize a single, end-to-end trainable architecture.


\subsection{Exploring Correspondences between Images}

Finding dense correspondences between a pair of images is one of the fundamental tasks of computer vision.
Input pairs can be different views of a scene \cite{hosni_fast_2013, hirschmuller_stereo_2008}, consecutive video frames \cite{horn_determining_nodate, teed_raft_2020} or even an image pair depicting different instances of the same semantic class as in the case of semantic correspondence matching \cite{liu_sift_nodate, rocco_convolutional_2017, lee_sfnet_2019}.
We aim to segment moving objects from otherwise static background; hence we focus on the second category where change detection \cite{radke_image_2005} or background modeling \cite{bouwmans_subspace_2009} are fundamental algorithms.
Due to superior performance of neural networks in many image-related tasks, \emph{Convolutional Neural Networks} (\emph{CNNs}) have also been applied~\cite{bouwmans_deep_2019}. De~Jong~et~al.~\cite{de_jong_unsupervised_2019} target the problem of unsupervised change detection in satellite images and propose a siamese architecture with feature fusion by concatenation. Ru~et~al.~\cite{ru_multi-temporal_2020} merge CNN features with a canonical correlation analysis to detect semantic changes in urban regions.

Optical flow is another way of obtaining pixel-wise correspondences between images, and current state-of-the-art approaches usually employ correlation in deep feature space to obtain a dense flow map (e.g. \cite{fischer_flownet_2015, hui_liteflownet_2018, teed_raft_2020}).

Since the rise of attention \cite{vaswani_attention_2017} several works explore differences between two data features with so-called \emph{co-attention}. 
This has initially been studied for vision-language tasks \cite{lu_hierarchical_2016, wang_learning_nodate}, but already has been applied successfully to the image domain: 
Lu et al. \cite{lu_see_2019} use co-attention for semantic video segmentation and achieve state-of-the-art performance for unsupervised video segmentation on DAVIS~\cite{perazzi_benchmark_2016} at their time of release. 




\subsection{Combination of Motion and Segmentation}
\label{sec:rl_combination}
Many dense correspondence estimators have successfully been coupled with specific semantic knowledge.
For instance, Cioppa~et~al.~\cite{cioppa_real-time_2020} propose a background modeling algorithm that is updated with predictions of a semantic segmentation network which has learned important object-related features and aids the overall pipeline.
Similar work exist for optical flow estimation: \cite{sevilla-lara_optical_2016} apply prior knowledge on a pixel's motion by defining different motion models for specific semantic regions. 
Bai~et~al.~\cite{bai_exploiting_2016} segment potentially moving objects, estimate flow individually and merge the predictions with background motion to improve accuracy.
In~\cite{bideau_best_2018}, classical geometric knowledge of moving objects is combined with high-level semantic image understanding in order to improve performance.
Hur~et~al.~\cite{hur_joint_2016} utilize flow to enforce temporal segmentation consistency, and simultaneously pose epipolar constraints inferred from semantic information on the motion estimation. 
Cheng~et~al.~\cite{cheng_segflow_2017} propose to jointly predict optical flow and semantic segmentation masks by employing a two stream approach, while \cite{tokmakov2019learning, jain2017fusionseg} segment moving objects by employing a specific motion and appearance network.

Nevertheless, the current best performing methods on the Kitti Optical Flow Benchmark \cite{noauthor_kitti_nodate} solely estimate optical flow. 
Concurrently we observe that flow architectures usually differ from semantic segmentation networks, as features for each task are extracted differently.
Moreover, we argue that optical flow itself over-complicates moving foreground extraction for object segmentation: for a respective binary mask it is solely interesting whether an object moves or not; direction and magnitude of each pixel do not yield any benefits.


Consequently, we rethink the utilization of motion in our self-supervised object segmentation pipeline. 
We propose an architecture that can efficiently estimate dense correspondences between features, and is additionally capable of incorporating specific semantic attributes.
In contrast to previous self-supervised grasped object segmentation approaches our framework consists of a single network that is trainable end-to-end, and does not require any post-processing.
\begin{figure*}[!ht]
	\vspace{3mm}
	\centering
	\captionsetup{width=\linewidth}
	\includegraphics[width=0.98\linewidth]{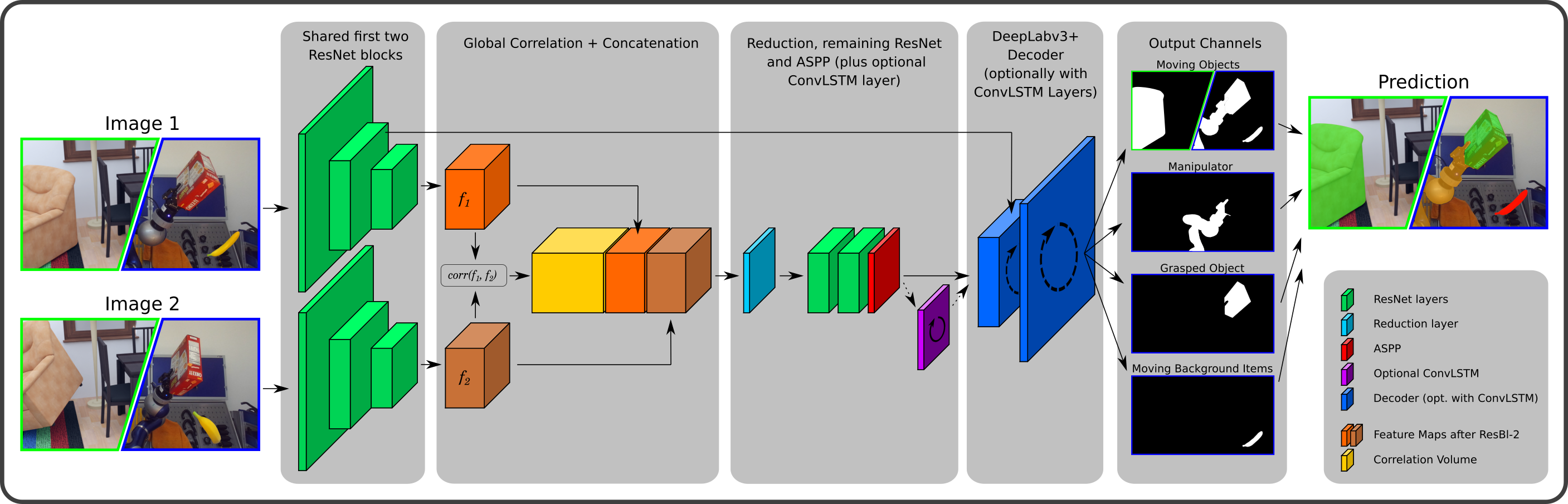}
	\caption{Architectural overview (best viewed magnified). \emph{DistinctNet} can be trained on moving foreground segmentation  (left image cuts; green border) and afterwards fine-tuned on semantically distinguishing foreground into manipulator, object and optional other semantic classes such as moving background objects (right image cuts; blue border). Dashed lines denote optional recurrent layers. Note that the last three output channels (here depicting manipulator, object and a moving background object) are only available during fine-tuning.}
	\label{fig:arch}
\end{figure*}%

\section{Method}



We modify a standard DeepLabv3+~\cite{chen_encoder-decoder_2018} with a \mbox{ResNet-50}~\cite{he_deep_2015} backbone which has proven to be successful for many semantic segmentation applications. 
In the following we refer to our network as \emph{DistinctNet}, as it allows to (1) distinguish moving foreground from static background as well as (2) semantically differentiate objects.
We now further explain our modifications that let us predict moving foreground and allow for fine-tuning on robot-object segmentation.
For an overview of our pipeline please see~Fig.~\ref{fig:arch}.

\subsection{Architecture}
\label{sec:architecture}
DistinctNet takes as input two images depicting the displacement of an object and otherwise static background.
This input pair is processed by the first two ResNet blocks in siamese fashion with shared weights.
A correlation layer then calculates the similarity between the two feature volumes to derive information on possible displacement, which is commonly done in disparity or optical flow estimation architectures (e.g. \cite{dosovitskiy2015flownet, ferrari_segstereo_2018}). 
Inspired by \cite{rocco_convolutional_2017} we normalize features before and after the correlation operation, and finally apply a \emph{ReLU} non-linearity.
The output thereof is concatenated with the two feature maps and channel-wise reduced
to fit the input dimension of the following ResNet block. 
Finally, forwarding through the remaining DeepLabv3+ modules generates pixel-wise probabilities of foreground and background.
Additional output channels concatenated to the final layer of the network are utilized to predict different parts of the moving foreground like manipulator and object.

Furthermore, we experiment with \emph{Convolutional LSTM} (\emph{ConvLSTM}) layers \cite{shi_convolutional_nodate} after the ASPP module and in the Decoder which let us predict object and manipulator even if their motion between consecutive frames comes to a halt.
We investigate the performance of these recurrent layers in Sec.~\ref{sec:extreme_stop}.
Last but not least, we evaluate the placement of the correlation operation in our network in Sec. \ref{sec:ablation}, where we also consider employing a co-attention layer~\cite{lu_hierarchical_2016,lu_see_2019}.



\subsection{Moving Object Segmentation}
\label{sec:moving_obj_seg}

We first train DistinctNet to learn a semantic agnostic representation of a moving object in front of static background. 
While there exist several segmentation datasets with moving items in a scene \cite{wang_cdnet_2014, perazzi_benchmark_2016, abu_alhaija_augmented_2018}, they are either limited to specific items, have multiple moving objects or additionally move the camera.
Consequently, we use BlenderProc~\cite{denninger_blenderproc_2019} to generate realistic-looking synthetic image pairs of static scenes with one random object in motion between two consecutive frames, which we explain in full detail in \mbox{Sec.~\ref{sec:dataset_fg}}.
The synthetic image pairs let the network effectively reason about an unknown moving object in an otherwise static scene.

\subsection{Distinction between Manipulator and Grasped Object}

After training moving foreground estimation, DistinctNet is well suited to distinguish a moving manipulator without any grasped object from static background. 
Following the self-supervised paradigm these automatically generated masks are further used as training data, and we explain the respective details in Sec. \ref{sec:dataset_robobj}.
Importantly, the architecture of DistinctNet directly allows for fine-tuning on object-manipulator differentiation by increasing the number of output channels in the final layer.
Now, the network is capable of moving foreground segmentation as well as semantically interpreting and distinguishing moving parts into manipulator and object in the additional channels.






The semantic foreground separation can be pushed even further: 
Remarkably, it allows us to resolve the constraint on static background as we can denote an additional channel for all moving items other than the grasped object and manipulator.
Our aim here is to enhance the practical applicability of our approach, given that a completely static background cannot always be assured.
An experimental evaluation of this is given in Sec.~\ref{sec:garbage}.

\section{Datasets and Implementation Details}
\subsection{Training Datasets}
\subsubsection{Moving Object Segmentation}
\label{sec:dataset_fg}
With BlenderProc \cite{denninger_blenderproc_2019} we sample a camera position inside a random room of the SunCG dataset \cite{song_semantic_2016}, select an object inside the camera's view and render ten frames while we randomly shift and rotate the respective item. 
We repeat this procedure to obtain 5,000 sequences of random moving objects in a static scene.
DistinctNet is then trained on moving object segmentation (90/10 train/val split), where an input pair consists of two images of one sequence.

\subsubsection{Fine-Tuning on Robot-Object Segmentation}
\label{sec:dataset_robobj}
The training data for this task should consist of moving robotic arms with a grasped object in front of random, stationary background.
Hence, for each training pair we duplicate a random image of the MS COCO dataset \cite{lin_microsoft_2015} to emulate diverse static background. 
We further sample two random robot arms (one for each image) and augment the presence of an object at the gripper's end effector with an occluding object.
As in our previous work \cite{boerdijk_self-supervised_2020} we derive the corresponding object pasting spots by opening and closing the gripper at static position and once again segment moving foreground with DistinctNet.
As occluders we use random objects from the ShapeNet dataset \cite{chang_shapenet_2015} whose orientation, texture and lighting can be realistically simulated with BlenderProc.
Note that while we use the same object for both images in a training pair its orientation can be different.
In total we generate 50,000 sample pairs (90/10 train/val split).

During training we separately augment each image with Gaussian noise, and add random color jitter for both the whole image as well as the occluding object. 
Additionally, to cope with pasting artifacts which can present a simple shortcut to the network during learning \cite{dwibedi_cut_2017}, we apply a median blur on the full image.

For our validation set we do not cut out robot masks, but directly paste objects on the test backgrounds.
We do not augment these images to verify the generalization capability onto novel environments and objects.

\subsection{Implementation Details}
\label{sec:implementation_details}
The global correlation layer restricts us to a fixed input size, and we consequently resize images to 414x746~pixels.
All experiments are conducted on a single Nvidia GeForce RTX 2080 Ti. 
We use AdamW \cite{loshchilov_decoupled_2019} as optimizer with a weight decay of 0.01.
For moving object segmentation we use standard cross entropy loss and set the learning rate to 1e-3 with a reduction to 1e-4 for all ResNet layers.
When fine-tuning on semantic classes we freeze all encoder weights before the correlation operation and train with a multi-class binary cross entropy loss combined with a reduced learning rate of 1e-4 for all trainable parameters.
Training takes roughly four (moving object segmentation) and three (semantic finetuning) days with a batch size of 2 on a single Nvidia GeForce RTX 2080.

\subsection{Recording Setup and Test Dataset}
\label{sec:dataset_test}
Our robot arm is a KUKA LBR4+ 
on a linear axis employed with a Robotiq 2F-85 two-finger gripper. 
All data is recorded with a ZED stereo sensor. 


We use the same 15 YCB object recordings as in our previous work (301 images per object), and refer the reader to~\cite{boerdijk_self-supervised_2020} for a detailed recording protocol.
45 images per object already contain manually annotated segmentation masks, and we additionally label the respective robot arms for a complete evaluation of foreground, robot and object.

\section{Experimental Evaluation}

In this section we evaluate the performance of our method on self-supervised grasped object segmentation. 
We further provide insights on the model's capability to deal with moving background objects and static input frames.
We continue by exploring different architectural settings to explain our network design choices.
Finally we show that the generated masks show equal or better performance for the task of semantic segmentation from scenes in comparison to a subset of ground truth annotations.


\subsection{Evaluation of Self-Supervised Object Segmentation}

We quantitatively evaluate our method in Tab. \ref{tab:quant_results} and report the \emph{mean Intersection over Union} (\emph{mIoU}) for all objects as well as robot arm, moving foreground and background.
Adding the gripper as separate semantic label (\mbox{Sec. \ref{sec:dataset_robobj}}) increases overall object mIoU by one percent point.

\begin{figure*}[!ht]
	\vspace{4mm}
	\centering
	\captionsetup{width=\linewidth}
	\includegraphics[width=0.9\linewidth]{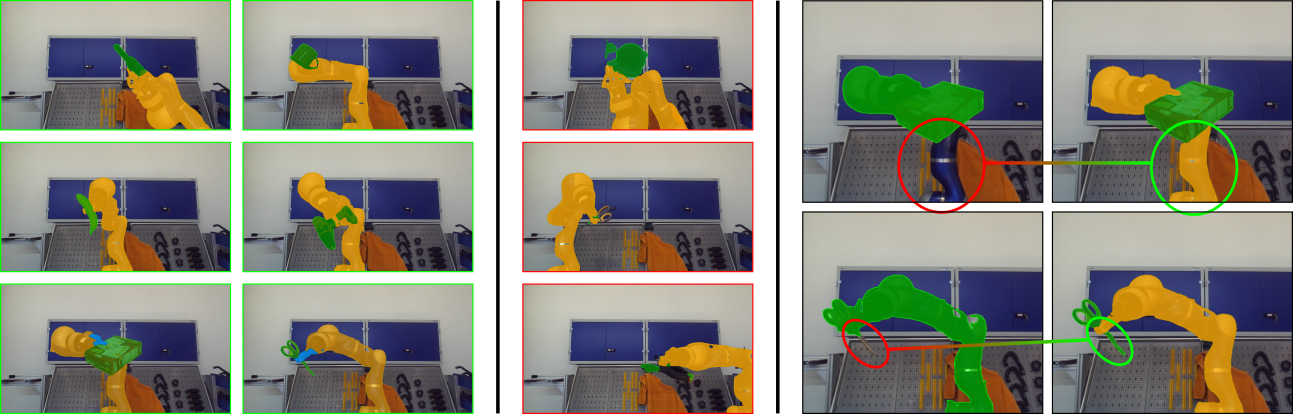}
	\caption{DistinctNet generates semantic annotations of a moving manipulator with a grasped object and optional other classes like the gripper (left images);
		Some items are particularly challenging because of similar color to surroundings or due to their shape (middle images);
		Fine-tuning on semantics increases performance on moving foreground extraction (right images).}
	\label{fig:qual_results}
\end{figure*}

DistinctNet generates strong segmentation masks (left part of Fig. \ref{fig:qual_results}), and we outperform recent self-supervised approaches with similar weak constraints on the setup by a large margin.
Despite not using camera calibration, manipulator key point registration and additional depth data like~\cite{florence_robot-supervised_2020} we still reach similar performance on a joint subset of objects (82.72\% vs. 84.61\% mIoU).

Nevertheless, some object views are quite challenging to our approach (middle part of Fig. \ref{fig:qual_results}), and a particularly outstanding item is the \emph{037\_scissors}. 
We hypothesize that due to the thin shape of the object the network finds it difficult to establish correspondences between two consecutive frames and identify it as moving object.
This would also explain the superior performance of \cite{florence_robot-supervised_2020} on this item, as their approach does not rely on identification by motion.

We concurrently observe that fine-tuning on semantics improves moving foreground extraction (right part of Fig.~\ref{fig:qual_results}):
Initially, DistinctNet segments joint robot-object masks from our test set with a mIoU score of 86.56\%. As reported in Tab. \ref{tab:quant_results} this increases by roughly six percent points after fine-tuning, indicating that semantic knowledge (albeit here merely in the form of background, object and manipulator) is important for segmentation from motion, which is in line with previous research (Sec.~\ref{sec:rl_combination}).
Furthermore, while we hypothesize that the correlation operation itself ensures strong generalization from synthetic to real data, the semantic fine-tuning might also be helpful for bridging a remaining sim2real gap.

\begin{table}[!hb]
	\centering
	\caption[Comparison to Baselines]{Mean IoU ([\%]) of our method compared to change detection and two recent approaches from literature.
\emph{Ours+G.} denotes predictions with an additional gripper output channel.
Numbers in \textbf{bold} depict best results.}
	\vspace*{1mm}
	\resizebox{0.95\columnwidth}{!}{
		\begin{tabular}{lccccc}
\toprule
Semantic Class&CD&\cite{florence_robot-supervised_2020}&\cite{boerdijk_self-supervised_2020}&Ours&Ours+G.\\
\midrule
\textbf{Object} & 33.24 & - & 75.80 & 83.11 & \textbf{84.21}\\
\hspace{3mm}003\_cracker\_box & 62.58 & - & 88.84 & 90.57 & \textbf{92.70}\\
\hspace{3mm}005\_tomato\_soup\_can & 24.86 & - & 81.53 & 85.95 & \textbf{87.66}\\
\hspace{3mm}006\_mustard\_bottle & 47.61 & - & 86.24 & 87.21 & \textbf{89.14}\\
\hspace{3mm}007\_tuna\_fish\_can & 13.95 & - & 60.81 & \textbf{75.44} & 74.89\\
\hspace{3mm}008\_pudding\_box & 32.34 & - & 80.40 & 90.28 & \textbf{91.74}\\
\hspace{3mm}010\_potted\_meat\_can & 26.18 & - & 81.78 & \textbf{85.39} & 84.68\\
\hspace{3mm}011\_banana & 28.54 & 65.40 & 77.23 & 83.14 & \textbf{86.36}\\
\hspace{3mm}019\_pitcher\_base & 32.84 & \textbf{95.70} & 78.56 & 89.47 & 89.43\\
\hspace{3mm}021\_bleach\_cleanser & 59.32 & - & 83.82 & 85.95 & \textbf{88.95}\\
\hspace{3mm}024\_bowl & 49.75 & \textbf{95.10} & 91.19 & 93.21 & 92.60\\
\hspace{3mm}025\_mug & 33.16 & \textbf{93.10} & 86.36 & 91.93 & 92.36\\
\hspace{3mm}035\_power\_drill & 35.06 & \textbf{84.40} & 60.69 & 74.72 & 75.86\\
\hspace{3mm}037\_scissors & 7.670 & \textbf{72.10} & 46.01 & 48.18 & 49.88\\
\hspace{3mm}052\_extra\_large\_clamp & 23.35 & - & 53.48 & 72.88 & \textbf{74.35}\\
\hspace{3mm}061\_foam\_brick & 21.33 & 86.50 & 80.08 & 92.32 & \textbf{92.53}\\
\textbf{Robot} & 54.07 & - & - & \textbf{91.42} & 91.39\\
\textbf{Foreground} & 52.02 & - & - & \textbf{92.19} & 92.11\\
\textbf{Background} & 93.82 & - & - & \textbf{99.08} & 99.07\\
\bottomrule
\end{tabular}}
	\label{tab:quant_results}
\end{table}

\subsection{Extreme Cases of Motion}
Our method typically relies on motion of manipulator and object.
Nevertheless, small modifications to our framework allow to deal with challenging scenarios such as moving background items or no observable motion at all.


\subsubsection{Dealing with Non-Static Background Objects}
\label{sec:garbage}
\begin{figure*}[!ht]
	\centering
	\subfloat[][Image 1]{\includegraphics[width=0.18\textwidth]{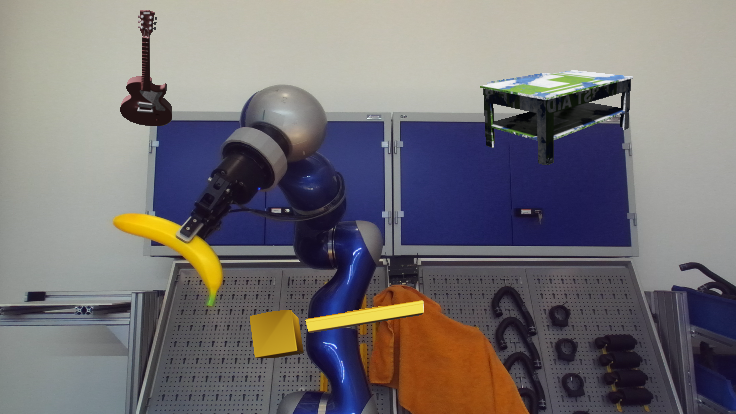}} 
	\enskip
	\subfloat[][Image 2]{\includegraphics[width=0.18\textwidth]{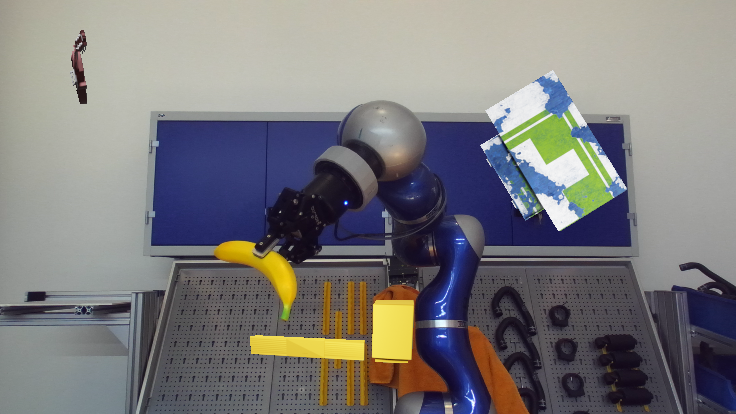}}
	\enskip
	\subfloat[][Prediction]{\includegraphics[width=0.18\textwidth]{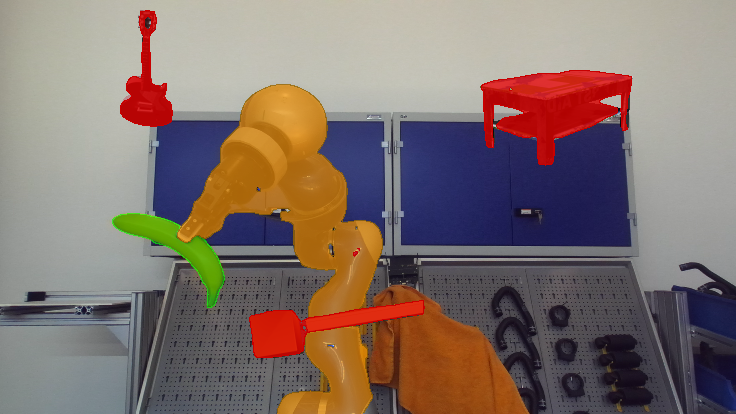}}
	\enskip
	\subfloat[][Object Heatmap]{\includegraphics[width=0.18\textwidth]{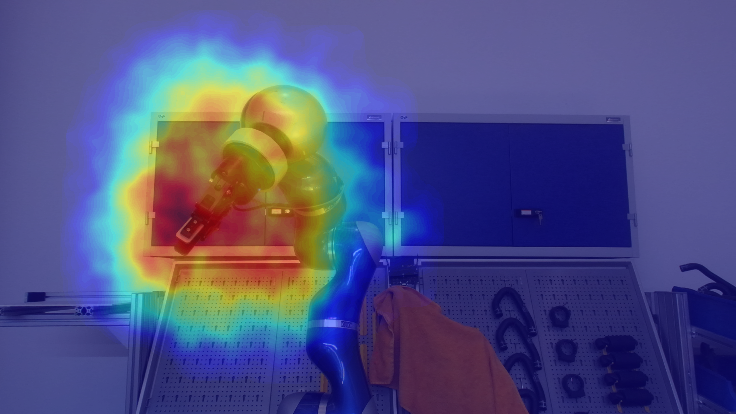}}	
	\enskip
	\subfloat[][Thresholded Flow]{\includegraphics[width=0.18\textwidth]{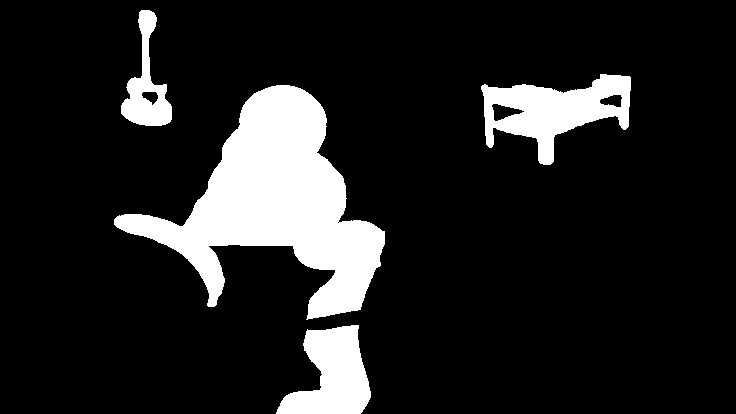}}
	
	\caption[Background motion]{Augmented moving background objects between two consecutive frames (a, b); Our network can directly separate moving foreground into grasped object (green), manipulator (orange) and other moving objects (red) (c); We sample 1,000 different objects on random positions on the image and visualize whether the item is identified as grasped object or belongs to moving background objects (d); While thresholded optical flow successfully segments moving foreground the binary prediction requires additional reasoning to differentiate objects into manipulator, object and distractor (e).}
	
	\label{fig:moving_background}
\end{figure*}

We can efficiently learn the presence of other moving objects by treating them as a new semantic class and adding another output channel to our network during semantic fine-tuning.
We emulate random background object motion by pasting additional objects onto our training data with different random spatial positions between consecutive frames.
After training, DistinctNet successfully differentiates a grasped object from other moving distractor items, as visualized in Fig. \ref{fig:moving_background}. Thereby, the overall object mIoU is merely reduced by 3\% due to mismatches between the grasped item and an object moving in the background.
This makes our approach more versatile in practical applications than our previous work, where every moving pixel not identified as manipulator is automatically classified as grasped object.

\subsubsection{Dealing with Motion Stops}
\label{sec:extreme_stop}
\begin{figure}[!b]
	\centering
	\captionsetup{width=\linewidth}
	\includegraphics[width=0.8\linewidth]{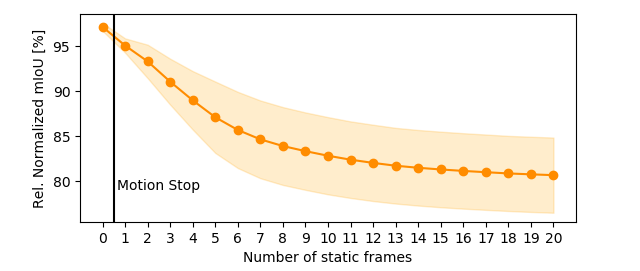}
	\caption{
		After ten consecutive frames we simulate a motion stop by feeding equal input pairs to the network
		(t=0 denotes the last moving input pair). 
		We repeat this five times per object, and plot mean and variance of the IoU relative to the point where motion last occurred across all objects.
		On average, DistinctNet still predicts 80\% of the initial segmentation mask after 20 static frames.}
	\label{fig:recurrent}
\end{figure}

To alleviate the constant dependency on motion for successful prediction we propose to add a \emph{ConvLSTM} layer after the ASPP module and replace every \emph{Conv2d} layer in the decoder with the respective recurrent counterpart (see also Fig. \ref{fig:arch}).
Whenever both input images are alike (i.e., no observable motion between frames), the network still estimates manipulator and object given the memorized predictions of previous frames.
We quantitatively analyze the decrease in mIoU after a motion stop in Fig. \ref{fig:recurrent}.


\subsection{Ablation Study}
\label{sec:ablation}
We experiment with various placements of the correlation layer and a co-attention operation \cite{lu_hierarchical_2016, lu_see_2019} to explain our architectural design choices. 
In \cite{lu_see_2019} co-attention is performed after the ASPP module, and we add an experiment with the co-attention block placed at the position of the best-performing correlation setting for fair comparison.
In Tab.~\ref{tab:architecture_ablation} we list the mIoU for the task of agnostic segmentation from motion (Sec. \ref{sec:moving_obj_seg}) with our synthetic validation set \mbox{(Sec. \ref{sec:dataset_fg})} and joint robot-object masks as test~set~\mbox{(Sec. \ref{sec:dataset_test})} after five epochs of training.
We further show the number of parameters in million. 
For the inference estimate we average ten forward passes (single batch size, including image pre-processing) in evaluation mode.
Our proposed setting of employing a correlation layer after the second ResNet block results in best performance on both data splits, while simultaneously being substantially faster than all other settings.
We hypothesize that the comparatively bad performance of the co-attention layer stems from strong augmentations on both images.

\begin{table}[!t]
	\centering
	\caption[Quantitative results]{Architectural ablation study on different merge layers and positions. Our setting (\textbf{bold}) results in best performance for moving foreground segmentation in terms of mIoU on validation and test set while operating at~\texttildelow14~Hz.}
	\vspace*{1mm}
	\resizebox{\columnwidth}{!}{
		\begin{tabular}{llccc}
	\toprule
	Merge Type & Pos. After & mIoU V/T [\%] & MParams & Inf. [ms]\\	
	\midrule

	Correlation & ResBl-1&87.76 / 79.36&85.03&229.69\\
	\textbf{Correlation} & \textbf{ResBl-2}&89.15 / 82.16&64.75&71.12\\	
	Correlation & ResBl-3&88.74 / 81.99&93.88&102.55\\
	Correlation & ResBl-4&84.19 / 67.67&166.28&172.68\\

	Correlation & ASPP&74.85 / 60.98&51.96&98.88\vspace{1.5mm}\\

	Co-Attention & ResBl-2 &28.05 / 02.88 &45.33&73.77\\
	Co-Attention & ASPP&56.91 / 39.42&41.59&111.22\\
	Co-Attention-Orig\textsuperscript{*} & ASPP&50.61 / 20.02&41.59&115.75\\  


	\bottomrule

\end{tabular}}
	\begin{tabular}{l}
	\footnotesize{*Here, segmentation maps for both inputs are predicted as per \cite{lu_see_2019}.}
	\end{tabular}

	\label{tab:architecture_ablation}
\end{table}






\subsection{Semantic Segmentation as Downstream Task}
We evaluate the quality of our automatically generated object masks by employing them as training data for semantic segmentation.
Specifically, we are interested in the performance of these masks in contrast to manually annotated object masks (i.e., the ground truth annotations we compare to in Tab. \ref{tab:quant_results}), and the improvements to our previous work.

\begin{figure}[!b]
	\centering
	\captionsetup{width=\linewidth}
	\includegraphics[width=0.8\linewidth]{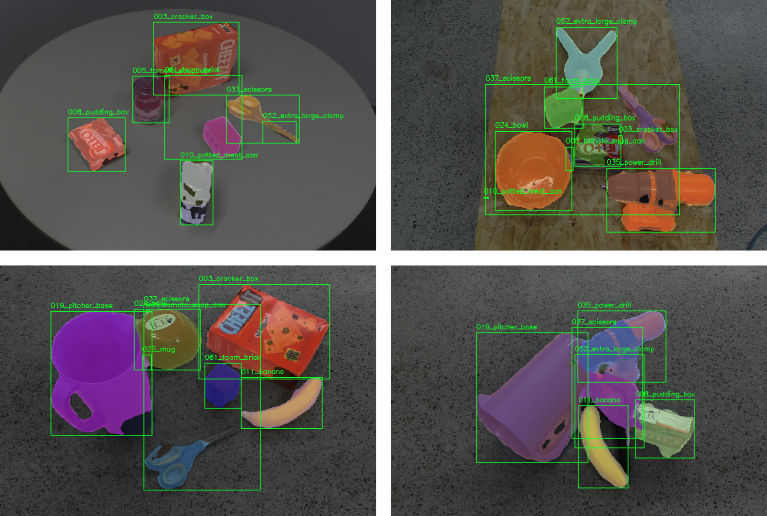}
	\caption{
		Qualitative results from our semantic segmentation from scenes (best viewed magnified). We add bounding boxes to denote the class labels of the respective annotations.}
	\label{fig:semseg}
\end{figure}

Diverse training data (50,000 images) is generated by pasting object masks
with random scale and translation on MS~COCO~\cite{lin_microsoft_2015} backgrounds. 
We train a standard DeepLabv3+ \cite{chen_encoder-decoder_2018} with the same settings as for moving object segmentation (Sec. \ref{sec:implementation_details}) and compare ground truth and predicted masks on images of STIOS \cite{durner2021unknown}
in Tab. \ref{tab:semseg}.
We visualize exemplary results in Fig. \ref{fig:semseg}.

For equal views across all masks the ground truth annotations represent an upper border, and we perform superior to masks generated by \cite{boerdijk_self-supervised_2020} due to better initial \mbox{quality (see Tab. \ref{tab:quant_results})}.
Remarkably, the network benefits from many additional views, and employing all object views segmented by our approach as training data performs better than the smaller subset of ground truth annotations.
Note that while there is no experiment with 301 ground truth masks per object we expect an analogous upper bound.

\begin{table}[t]
	\centering
	\caption[Quantitative results]{Mean IoU on segmentation from tabletop scenes. Training data are a subset (45) or all (301) masks per grasped object. We use our generated annotations, predictions from~\cite{boerdijk_self-supervised_2020} as well as ground truth annotations.}
	\vspace*{1mm}
	\resizebox{0.7\columnwidth}{!}{
		\begin{tabular}{lccc}
\toprule

\#Masks / Obj. & Ground Truth & \cite{boerdijk_self-supervised_2020} & Ours\\
\midrule
45 Masks& 58.91 & 50.39 &56.79\\
301 Masks& - & 54.78 &59.86\\
\bottomrule


\end{tabular}

}
	\label{tab:semseg}
\end{table}

\section{Conclusion}
We have presented a novel method for self-supervised grasped object segmentation from robotic manipulation.
Although all our learning is based on observing motion in a static scene, a major contribution of this work is the identification and distinction of other moving background objects, as well as coping with stagnating motion.
Our framework presents a reliable way of large-scale training data generation from real sensor environments, and the resulting masks are well suited for downstream tasks like semantic segmentation.

\clearpage
\bibliographystyle{IEEEtran}
\bibliography{refs,ext_refs} 

\end{document}